\documentclass[conference]{IEEEtran}
\usepackage{times}
\usepackage{amsmath}
\usepackage{amsfonts}
\usepackage{amssymb}
\usepackage{graphicx}
\usepackage{float}
\usepackage{stfloats}
\usepackage[numbers]{natbib}
\usepackage{multicol}
\usepackage{multirow}
\usepackage[bookmarks=true]{hyperref}
\usepackage{nicefrac}       % compact symbols for 1/2, etc.
\usepackage{microtype}      % microtypography
\usepackage{algorithm}
\usepackage[noend]{algpseudocode}
\usepackage{xcolor}

\algnewcommand{\LineComment}[1]{\State \(\triangleright\) #1}
\usepackage{todonotes}

\newcommand{\envembedding}{h}
\newcommand{\encoderparams}{\theta}
\newcommand{\encoder}{q_\encoderparams(\envembedding|E)}
\newcommand{\decoderparams}{\psi}
\newcommand{\decoder}{p_\decoderparams(E|h)}
\newcommand{\vaepriorparams}{\phi}
\newcommand{\vaeprior}{p_\vaepriorparams(h)}

\newcommand{\context}{C}
\newcommand{\contextnetparams}{\omega}
\newcommand{\contextnet}{g_\contextnetparams}
\newcommand{\flowparams}{\zeta}
\newcommand{\flow}{f_\flowparams}

\newcommand{\cost}{J}

\begin{document}

% paper title
\title{Variational Inference MPC using Normalizing Flows and Out-of-Distribution Projection}

% You will get a Paper-ID when submitting a pdf file to the conference system
\author{Author Names Omitted for Anonymous Review. Paper-ID 186}

\author{\authorblockN{Thomas Power and Dmitry Berenson}
\authorblockA{Robotics Institute, University of Michigan, Ann Arbor, MI 48109\\
Email: \{tpower, dmitryb\}@umich.edu}
}

\maketitle

\begin{abstract}
We propose a Model Predictive Control (MPC) method for collision-free navigation that uses amortized variational inference to approximate the distribution of optimal control sequences by training a normalizing flow conditioned on the start, goal and environment. This representation allows us to learn a distribution that accounts for both the dynamics of the robot and complex obstacle geometries. We can then sample from this distribution to produce control sequences which are likely to be both goal-directed and collision-free as part of our proposed FlowMPPI sampling-based MPC method. However, when deploying this method, the robot may encounter an out-of-distribution (OOD) environment, i.e. one which is radically different from those used in training. In such cases, the learned flow cannot be trusted to produce low-cost control sequences. To generalize our method to OOD environments we also present an approach that performs \textit{projection} on the representation of the environment as part of the MPC process. This projection changes the environment representation to be more in-distribution while also optimizing trajectory quality in the true environment. Our simulation results on a 2D double-integrator and a 3D 12DoF underactuated quadrotor suggest that FlowMPPI with projection outperforms state-of-the-art MPC baselines on both in-distribution and OOD environments, including OOD environments generated from real-world data.
%\todo{update if including arms}
\end{abstract}

\IEEEpeerreviewmaketitle

\section{Introduction}
\label{sec:intro}
Model predictive control (MPC) methods have been widely used in robotics for applications such as autonomous driving \cite{mppi}, bipedal locomotion \cite{mpc_bipeds} and manipulation of deformable objects \cite{manipulation_mppi}. For nonlinear systems, sampling based approaches for MPC such as the Cross Entropy Method (CEM) and Model Predictive Path Integral Control (MPPI) \cite{CEM, mppi} have proven popular due to their ability to handle uncertainty, their minimal assumptions on the dynamics and cost function, and their parallelizable sampling. However, these methods struggle when randomly-sampling low-cost control sequences is unlikely and can become stuck in local minima, for example when a robot must find a path through a cluttered environment. This problem arises because the sampling distributions used by these methods are not informed by the geometry of the environment.
\begin{figure}
    \centering
    \includegraphics[width=0.48\textwidth]{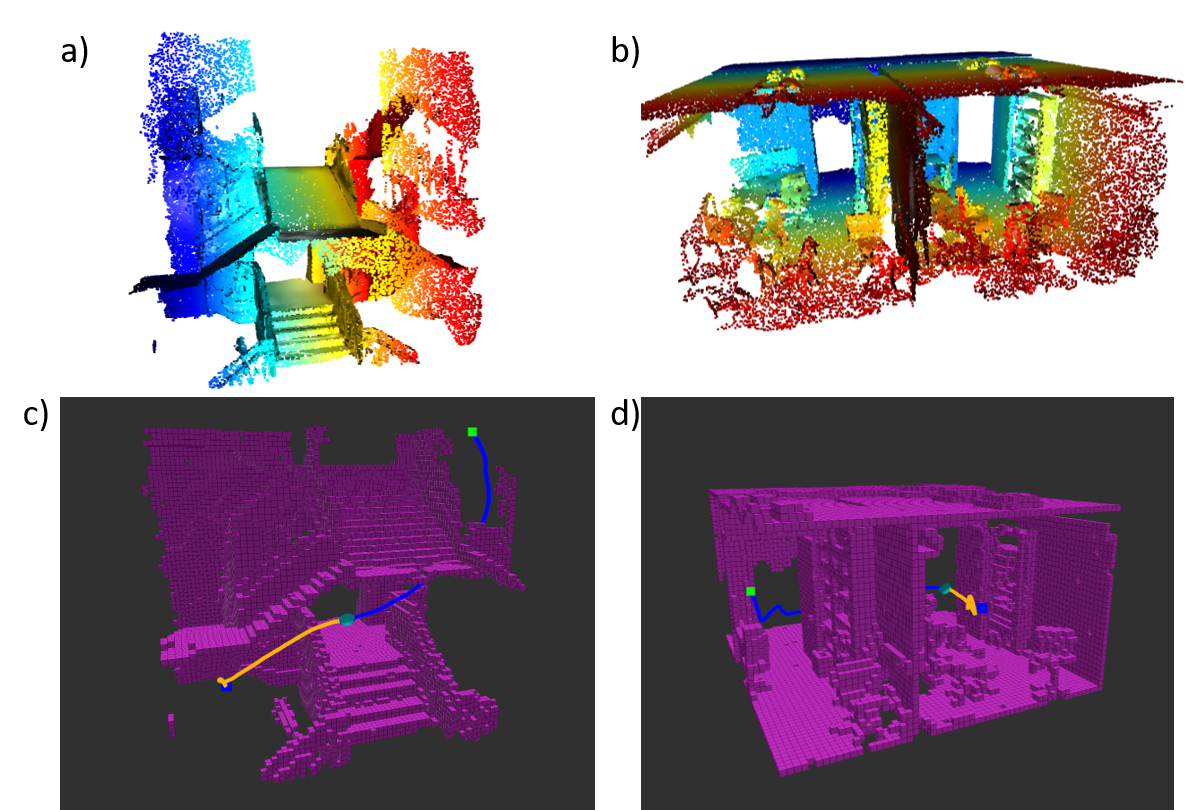}
    \caption{a,b) Point clouds of two real-world environments taken from the 2D-3D-S dataset \cite{3d_dataset}. c,d) Our method, FlowMPPIProject, controlling a dynamic quadcopter to successfully traverse these two environments. The executed trajectory is shown in blue, and the planned trajectory is shown in orange at an intermediate point in the execution}
    \label{fig:realworld}
\end{figure}

Previous work has investigated the duality between control and inference \cite{todorov_duality_2008, todorov_duality_2} and considered both planning and control as inference problems \cite{planning_as_inference, toussaint_probabilistic_2006, soc_as_inference}. Several recent papers have considered the finite-horizon stochastic optimal control problem as Bayesian inference, and proposed methods of performing variational inference to approximate the distribution used to sample control sequences \cite{stein_mpc, VI_Tsallis, okada_variational_mpc, dual_stein}. In order to perform variational inference, we must specify a parameterized distribution which is tractable to optimize and sample while also being flexible enough to provide a good approximation of the true distribution over low-cost trajectories, which may exhibit strong environment-dependencies and multimodalities. While more complex representations have been used to represent this distribution \cite{stein_mpc,okada_variational_mpc}, these distributions are initially uninformed and must be iteratively improved during deployment. Instead, our proposed method uses a normalizing flow to represent this distribution and we learn the parameters for this model from data. %To our knowledge, this paper presents the first method that learns a control sequence sampling distribution for MPC from a dataset. 

The advantage of this approach is that it will learn to sample control sequences which are likely to be both goal-directed and collision-free (i.e. low-cost) for the given system. We use the learned distribution as part of our proposed FlowMPPI sampling-based MPC method. This method samples perturbations to a nominal trajectory in \textit{both} the latent space of the flow and the space of control sequences.

However, as is common in machine learning, a learned model cannot be expected to produce reliable results when its input is radically different from the training data. Because the space of possible environments is very high-dimensional, we cannot hope to generate enough training data to cover the set of possible environments a robot could encounter. This problem compounds when we generate training data in simulation, but the method must be deployed in the real-world (i.e. the sim2real problem). Thus, when deploying this method, the robot may encounter an out-of-distribution (OOD) environment, i.e. one which is radically different from those used in training. In such cases, the learned distribution in unlikely to produce low-cost control sequences. 

To generalize our method to OOD environments we present an approach that performs \textit{projection} on the representation of the environment as part of the MPC process. This projection changes the environment representation to be more in-distribution while also optimizing trajectory quality in the true environment. In essence, this method ``hallucinates'' an environment that is more familiar to the normalizing flow so that the flow produces reliable results. However, the key insight behind our projection method is that the ``hallucinated'' environment cannot be arbitrary, it should be constrained to preserve important features of the true environment for the MPC problem at hand. For example, consider a navigation problem for a 2D point robot, shown in Figure \ref{fig:projection}. If the normalizing flow is trained only on environments consisting of disc-shaped obstacles, an environment with a corridor would be OOD and the flow would be unlikely to produce low-cost trajectories. However, if we morph the environment to approximate the corridor near the robot with disc-shaped obstacles (producing an in-distribution environment), the flow will then produce low-cost samples for MPC.

Our simulation results on a 2D double-integrator and a 3D 12DoF underactuated quadrotor suggest that FlowMPPI with projection outperforms state-of-the-art MPC baselines on both in-distribution and OOD environments, including OOD environments generated from real-world data (Figure \ref{fig:realworld}). %In addition, FlowMPPI outperforms these same baselines when evaluating on environments generated with real world data. 

The contributions of this paper are:
\begin{itemize}
    \item A method to learn an environment-dependent sampling-distribution of low-cost control sequences using a Normalizing Flow
    \item FlowMPPI - A method that computes a low-cost control sequence by sampling perturbations to a nominal control sequence in both the latent space of the learned normalizing flow and the space of control sequences
    \item A projection method which changes the environment representation to be more in-distribution while preserving important features of the environment for the MPC problem at hand
    \item Experiments showing the efficacy of our method on both in-distribution and OOD environments for planar navigation and 12DoF quadrotor tasks, including environments generated from real-world data 
\end{itemize}

%SVI-MPC approximates the posterior as a set of particles \cite{stein_mpc}, whereas  VI-MPC uses a mixture of Gaussians \cite{okada_variational_mpc}. We propose using a normalizing flow to approximate the true posterior over trajectories. Normalizing flows have previously been used to parameterize variational posteriors in \cite{pmlr-v37-rezende15}.
%\paragraph{Contributions:} We propose a method for learning a start, goal and environment conditioned distribution over finite horizon action sequences for MPC applied to navigation problems. Our method is able to generate diverse action sequences which display goal driven behavior for use downstream in a sampling-based MPC controller and does not require differentiable cost or dynamics. We also propose a method for generalizing this action sequence distribution to novel environments. We attempt to find an in-distribution environment from which we can sample high-performing action sequences evaluated in the actual novel environment. We demonstrate that our approach can generalize to difficult novel environments outside the training distribution where MPPI fails on a navigation problem. The outline of our approach is shown in Figure \ref{fig:framework}.

\section{Related Work}
\label{sec:lit}

\subsection{Planning \& Control as Inference}
The connection between control and inference is long established \cite{kalman1960, todorov_duality_2008, todorov_duality_2}.
\citet{planning_as_inference} first framed planning as an inference problem, and proposed a tractable inference algorithm for discrete state and action spaces. Further work has used inference techniques for planning \cite{GPmotion_planning, Gp_mp2} and Stochastic Optimal Control (SOC) \cite{toussaint_probabilistic_2006, soc_as_inference, watson_socinf}. Two widely used sampling based MPC techniques, MPPI \cite{mppi} and CEM \cite{CEM}, use importance sampling to generate low-cost control sequences, and have strong connections to the inference formulation of SOC which was explored in \cite{VI_Tsallis}. Several recent papers have considered the SOC problem as Bayesian inference, and proposed methods of performing Variational Inference (VI) to approximate a posterior over low-cost control sequences \cite{stein_mpc, VI_Tsallis, okada_variational_mpc, dual_stein}. These methods differ in how they represent the variational posterior. VI methods often use an independent Gaussian posterior, known as the mean-field approximation \cite{VI}.  \citet{okada_variational_mpc} represent the control sequence as a Gaussian mixture, and \citet{stein_mpc} use a particle representation, extended to handle parameter uncertainty in \cite{dual_stein}. These representations allow for greater flexibility in representing complex posteriors. We will similarly use a flexible class of distributions to represent the posterior, but will further make the posterior dependent on the start, goal, and environment. To our knowledge our approach is the first to amortize the cost of computing this posterior by learning a conditional control sequence posterior from a dataset.

\subsection{Learning sampling distributions for planning} Our work is related to work learning sampling distributions from data for motion planning. \citet{learned_sampling2} proposed learning a sampling distribution that is trained across multiple environments, but is independent of the environment. Others have proposed learning a sampling distribution which is dependent on the environment, start and goal \cite{learned_sampling, learned_sampling_iros}. These methods were restricted to geometric planning, but \citet{mpcmnet} proposed an approach for kinodynamic planning which learns a generator and discriminator which are used to sample states that are consistent with the dynamics. Recent work by \citet{learned_sampling_flow} uses a diffeomorphism to learn the sampling distribution; a model that is similar to a normalizing flow. The model we propose will also learn to generate samples conditioned on the start, goal and environment, though in this work we are considering online MPC and not offline planning. \citet{Loew-RSS-21} uses probabilistic movement primitives (ProMPs) learned from data as the sampling distribution for sample-based trajectory optimization, however the representation of these ProMPs only allows for uni-modal distributions and the sampling distribution is not dependent on the environment. Adaptive and learned importance samplers have been used for sample-based MPC \cite{adaptive_importance_sampling, learned_importance_sampling}, but these methods only consider a single control problem and the learned samplers do not generalize to different goals \& environments.

%\subsection{Out-of-distribution detection in generative models}
%\rev{I haven't decided what to do about the OOD stuff  needs more though}
%\cite{vae_ood1, vae_ood2}
%Previous work has demonstrated that the likelihood from generative models is not necessarily indicative of an input being OOD. In fact, many generative models, including flows, can assign higher likelihood to OOD data \cite{ood_gen_models1, ood_gen_models2, flows_ood}. However, it has been noted that representations which have been trained for a downstream task can yield more reliable OOD detection due to the presence of semantic information \cite{flows_ood}. By training our entire system end-to-end, $h$ is trained to contain an encoding of the environment relevant for trajectory generation

\section{Problem Statement}
\label{sec:problem_statement}
This paper focuses on the problem of Finite-horizon Stochastic Optimal Control. We consider a discrete-time system with state $x \in \mathbb{R}^{d_x}$ and control $u \in \mathbb{R}^{d_u}$ and known transition probability $p(x_{t+1}|x_t, u_t)$. We define finite horizon trajectories with horizon $T$ as $\tau = (X, U)$, where $X = \{x_0, x_1, ... x_T\}$ and $U = \{u_0, u_1, ... u_{T-1}\}$. 

Given an initial state $x_0$, a goal state $x_G$, and a signed-distance field (SDF) of the the environment $E$, our objective is to find $U$ which minimizes the expected cost $E_{p(X|U)}[\cost(\tau)]$ for a given cost function $\cost$, where $p(X|U) = \prod^{T-1}_{t=0}p(x_{t+1}|x_t, u_t)$. Note that we will use $\cost$ to mean both the cost on the total trajectory $\cost(\tau)$ and the cost of an individual state action pair $\cost(x, u)$. This paper focuses on the problem of collision-free navigation, where $\cost$ is parameterized by $(x_G, E)$. 

This problem is difficult to solve in the general case because the mapping from environments to collision-free $U$ can be very complex and depends on the dynamics of the system. To aid in finding $U$, we assume access to a dataset $\mathcal{D} = \{E, x_0, x_G\}^N$, which will be used to train our method for a given system. We will evaluate our method in terms of its ability to reach the goal without colliding and the cost of the executed trajectory. Moreover, we wish to solve this problem very quickly (i.e. inside a control loop), which limits the amount of computation that can be used.

\section{Preliminaries}

%\subsection{Finite-horizon Stochastic Optimal Control} 

\subsection{Variational Inference for Stochastic Optimal Control}
\label{background:vi_soc}
We can reformulate SOC as an inference problem (as in \cite{soc_as_inference, soc_as_inference_2, okada_variational_mpc, stein_mpc}). First, we introduce a binary `optimality' random variable $o$ for a trajectory such that

\begin{equation}
    p(o=1|\tau) \propto \exp{(-\cost(\tau))}
\end{equation}

We place a prior $p(U)$ on $U$, resulting in a prior on $\tau$, $p(\tau) = p(X|U)p(U)$ and aim to find posterior distribution $p(\tau | o=1) \propto p(o=1 | \tau) p(\tau)$.
In general, this posterior is intractable, so we use variational inference to approximate it with a tractable distribution $q(\tau)$ which minimizes the KL-divergence  $\mathcal{KL}(q(\tau) || p(\tau | o = 1))$ \cite{VI}. Since we define the trajectory by selecting the controls, the variational posterior factorizes as $p(X|U)q(U)$. Thus, we must compute an approximate posterior over control sequences. The quantity to be minimized is
\begin{align}
\begin{split}
&\mathcal{KL}\left(q (\tau) || p(\tau | o=1) \right) = \int q(\tau) \log \frac{q(\tau)}{p(\tau | o=1)} d\tau \\
&= \int q(X, U) \log \frac{p(X|U)q(U)p(o=1)}{p(o=1|X, U)p(X|U)p(U)}dXdU 
\end{split}
\end{align}
Simplifying and omitting terms that do not depend on $\tau$ yields the \textit{variational free energy}
\begin{equation}
\mathcal{F} = -\mathbb{E}_{q(\tau)}[\log p(o| \tau) + \log p(U)] - \mathcal{H}(q(U))
\label{background:free_energy}
\end{equation}
Where $\mathcal{H}(q(U)$ is the entropy of $q(U)$. Intuitively, we can understand that the first term promotes low-cost  trajectories, the second is a regularization on the control, and the entropy term prevents the variational posterior collapsing to a \textit{maximum a posteriori} (MAP) solution. Note that $\log p(U)$ can be appropriately combined with the cost, i.e. a Gaussian prior can be incorporated as a squared cost on the control, so will be omitted for the rest of the paper. 

\subsection{Variational Inference with Normalizing flows}
Normalizing flows are bijective transformations that can be used to transform a random variable from some base distribution (i.e. a Gaussian) to a more complex distribution \cite{pmlr-v37-rezende15, real-nvp, glow}. Consider a random variable $z \in \mathbb{R}^d$ and with known pdf $p(z)$. Let us define a bijective function $f: \mathbb{R}^d \to \mathbb{R}^d$ and a random variable $y$ such that $y = f(z)$ and $z = f^{-1}(y)$. According to the change of variable formula, we can define $p(y)$ in terms of $p(z)$ as follows:
\begin{align}
    p(x) &= p(z) \left| \det \frac{\partial f}{\partial z} \right|^{-1} \\
    \log p(y) &= \log p(z) - \log \left| \det \frac{\partial f}{\partial z} \right|
    \label{background:flow_ll}
\end{align}
Normalizing flows can be used as a parameterization of the variational posterior \cite{pmlr-v37-rezende15}. By selecting a base PDF $p(z)$ and a family of parameterized functions $f_\theta$, we specify a potentially complex set of possible densities $q_\theta(y)$. Suppose that we want to approximate some distribution $p(y)$ with some distribution $q_\theta(y)$. The variational objective is to minimize $\mathcal{KL}(q_\theta(y) || p(y))$. This is equivalent to:
\begin{align}
\begin{split}
    &\mathcal{KL}\left(q_\theta (y) || p(y) \right) = \int q_\theta(y) \log \frac{q_\theta(y)}{p(y)} dx \\
    &= \mathbb{E}_{q_\theta(y)} [\log q_\theta(y) -\log p(y)] \\
    &= \mathbb{E}_{p(z)} \left[ \log p(z) - \log \left| \det \frac{\partial f_\theta}{\partial z} \right| -\log p(y) \right]
\end{split}
\end{align}
Thus we can optimize the parameters $\theta$ of the bijective transform $f_\theta$ in order to minimize the variational objective. We will use a normalizing flow to represent the control sequence posterior in our method.

\section{Methods}
\label{sec:method}
% \begin{figure*}
%     \centering
%     \includegraphics[width=\textwidth]{figures/FullArchitecture.png}
%     \label{fig:framework}
%     \caption{The overall architecture for the flow sampler. a) The inputs to the system are a start, goal and SDF representation of the environment. b) The environment is encoded into an embedding $\envembedding$. During training, we can evaluate the $\mathbb{E}_{\encoder} [\encoder]$ term in eq. (\ref{eq:vae_loss}). c) The start and goal are combined with the $\envembedding$ and embedded into a context vector $\context$ with MLP $\contextnet$. d) The environment embedding $\envembedding$ is used to reconstruct the SDF with the decoder, and during training, the $\mathbb{E}_{\encoder} [\decoder - \vaeprior]$ terms in eq. (\ref{eq:vae_loss}) are evaluated. e) The context vector $\context$ is used to generate control sequence samples with flow $\flow$}
% \end{figure*}

\begin{figure*}
    \centering
    \includegraphics[width=\textwidth]{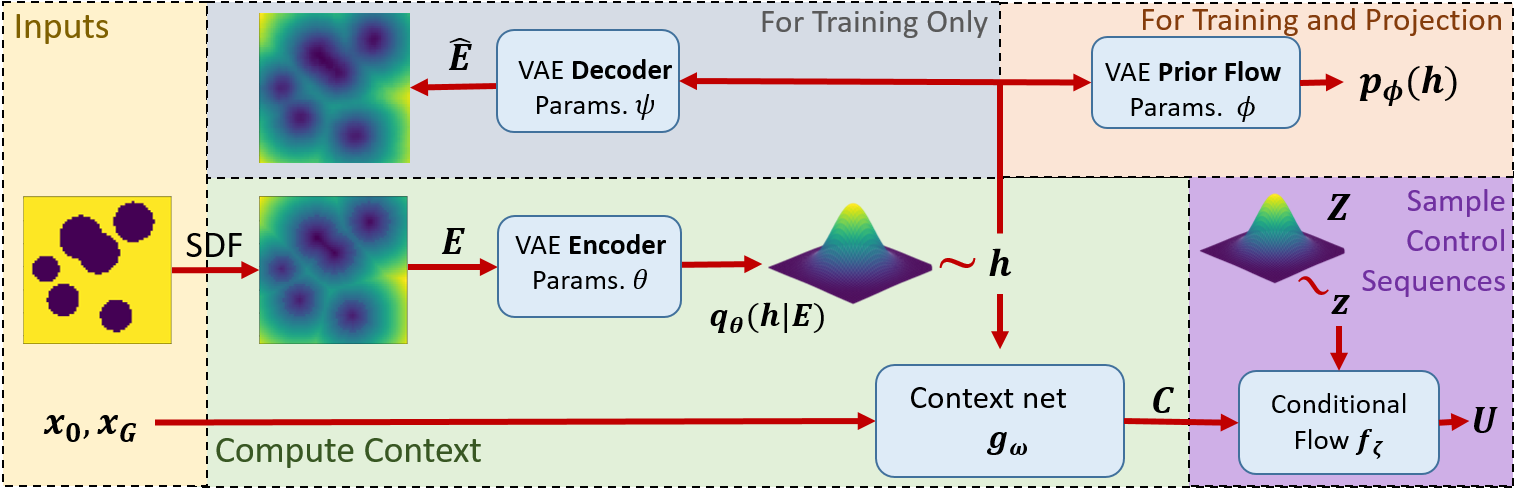}
    \caption{The architecture of our method for sampling control sequences. We take as input initial and goal states $x_0$, $x_G$, and the environment, converted to a signed distance field $E$.  $E$ is input into a VAE to produce a latent distribution $q_\theta(h|E)$, which we sample to get the environment embedding $h$. This $h$ is used, along with $x_0$ and $x_G$ as input to the network $g_\omega$ to produce a context vector $C$. $C$, along with a sample from a Gaussian distribution $Z$, is input into the conditional normalizing flow $\flow$ to produce a control sequence $U$. During training only, we use a decoder to reconstruct the SDF from $h$ as part of the loss. We also use a normalizing flow prior for the VAE to compute an OOD score for a given $h$, which is necessary to perform projection.}
    \label{fig:framework}
\end{figure*}

Our proposed architecture for learning an MPC sampling distribution is shown in Figure \ref{fig:framework}. In this section we first introduce how we represent and learn the control sequence posterior as a Normalizing Flow, and train over a dataset consisting of starts, goals and environments to produce a sampling distribution for control sequences. %Save it for later: Our method uses a VAE to encode an environment SDF into an environment embedding \todo{vector}. This environment embedding is used along with the start and goal as the input to a control sampling distribution, which samples control sequences. These control sequences produce goal-directed trajectories that avoid collision. 
Next, we show how this sampling distribution can be used to improve MPPI, a sampling based MPC controller. Finally, we describe an approach for adapting the learned sampling distribution to novel environments which are outside the training distribution. %save for later: To do this, we will use the environment embedding VAE to inform a differentiable out-of-distribution score, and use this top \textit{project} the environment embedding in distribution.

\subsection{Overview of Learning the Control Sequence Posterior} 

The control sequence posterior introduced in section \ref{background:vi_soc} is specific to each MPC problem. Our approach is to use dataset $\mathcal{D}$ to learn a conditional control sequence posterior $q(U|x_0, x_G, E)$. We will use a Conditional Normalizing Flow \cite{conditional_flow} to represent this conditional posterior as $q_\flowparams(U|\context)$. $\context$ is the context vector which we compute as follows: First, we input $E$ into the encoder of a Variational Autoencoder (VAE) \cite{vae} to produce a distribution over environment embeddings $\envembedding$. We then sample from this distribution to produce an $\envembedding$. A neural network $\contextnet$ then produces $\context$ from $(x_0, x_G, \envembedding)$ (Figure \ref{fig:framework}). Essentially $\context$ is a representation of what is important about the start, goal, and environment for generating low-cost trajectories.

%is computed by neural network $\contextnet$ from $(x_0, x_G, \envembedding)$, where $\envembedding$ is an environment embedding (Figure \ref{fig:framework}). This environment embedding is sampled from the latent space of a Variational Autoencoder (VAE) \cite{vae}. The VAE consists of an encoder which encodes $E$ into a distribution over $\envembedding$, a decoder which decodes $\envembedding$ to $\hat{E}$, and a learned prior $\vaeprior$, which is a normalizing flow. 

The above models are trained on the dataset $\mathcal{D}$,  which consists of randomly sampled starts, goals and simulated environments. To train the system we iteratively %we compute $(\envembedding, \context, \hat{E})$ and 
generate samples from the control sequence posterior, weigh them by their cost, and perform a gradient step on the parameters of our models to maximize the likelihood of low-cost trajectories. %. We then train with two losses: 1) A loss which maximizes the likelihood of sampled control sequences according to their cost (this loss does not require differentiable cost or dynamics functions); and 2) the standard VAE loss\todo{what about a loss from the VAE prior flow?}.

At inference time, we simply compute $\context$ and generate control sequence samples from $q_\flowparams(U|\context)$. Below we describe each component of the method to learn $q_\flowparams(U|\context)$ in detail.

%To generalize to OOD environments we apply a projection procedure based on an OOD score derived from $p_\vaepriorparams(\envembedding)$, to project $\envembedding$ to be more in-distribution when running our MPC algorithm. 

%The algorithm for projection is found in algorithm \ref{alg:train} in appendix \ref{appendix:algs}.

\subsection{Representing the start, goal and environment as $\context$}
\label{sec:vae}
As discussed, our dataset $\mathcal{D}$ consists of environments, starts and goals. The details of the dataset generation for each task can be found in section \ref{eval:tasks}. Since the environment is a high dimensional SDF, we must first compress it to make it computationally tractable to train the control sequence posterior. To encode the environment, we use a VAE with environment embedding $h$. The VAE consists of an encoder $\encoder$, which is a Convolutional Neural Network (CNN) that outputs the parameters of a Gaussian. The decoder is a transposed CNN which produces the reconstructed SDF $\hat{E}$ from $h$. The decoder log-likelihood $\decoder$ is $||\hat{E} - E||_2$, where $\decoderparams$ are the parameters of the decoder CNN. \citet{vae_flow_prior} showed that learning a latent prior can improve VAE performance, so we parameterize the latent prior $\vaeprior$ as a normalizing flow and learn the prior during training. The loss for the VAE is as follows:
\begin{align}
\begin{split}
    &\mathcal{L}_{VAE} = \mathbb{E}_{\encoder}\left[-\log \decoder \right] + \mathcal{KL}(\encoder || \vaeprior ) \\
    &= \mathbb{E}_{\encoder}\left[-\log \decoder + \log \encoder - \log \vaeprior) \right]
    \label{eq:vae_loss}
\end{split}
\end{align}
We then use a Multilayer Perceptron (MLP) network $\contextnet$ to generate a context vector $\context$ to use in the normalizing flow, via $\context = \contextnet(x_0, x_G, \envembedding)$, which has parameters $\contextnetparams$.

\subsection{Learning $q_\flowparams(U|C)$}

We use a conditional normalizing flow parameterized by $\flowparams$ to define the conditional variational posterior, i.e. $q_\flowparams(U | C)$ is defined by $U = \flow(Z, C)$ for $Z \sim p(Z) = \mathcal{N}(0, I)$. The variational free energy \ref{background:free_energy} then becomes:
\begin{align}
\begin{split}
    \mathcal{F} = -&\mathbb{E}_{q(\tau)} \left [\log p(o | \tau) \right] + \\ 
    &\mathbb{E}_{p(Z)} \left[ \log p(Z) -  \log \left |\det \frac{\partial f_\flowparams(Z, C)}{\partial Z} \right | \right] \label{eq:free_energy_flowmpc}
\end{split}
\end{align}
We can then optimize $\flowparams$ to minimize the free energy. 
% \begin{align}
% q^*_\zeta (U|C) = \arg \min_{q_{\flowparams}} \mathcal{F}
% \end{align}

By using a conditional normalizing flow, we are amortizing the cost of computing the posterior across environments. The conditional normalizing flow $U = f_\flowparams(Z, C)$ is invertible with respect to $Z$, i.e. $Z = f^{-1}(U, C)$. For our conditional Normalizing Flow we use an architecture based on Real-NVP \cite{real-nvp} architecture with conditional coupling layers \cite{conditional_flow}, the structure is specified in section \ref{eval:architectures}.

Minimizing eq. (\ref{eq:free_energy_flowmpc}) via gradient descent requires the cost and dynamics to be differentiable. To avoid this, we estimate gradients, using the method in \cite{okada_variational_mpc}: At each iteration, we sample $R$ control sequences $U_{1..R}$ from $q_\flowparams(U|C)$ and compute weights %For exploration, we apply a Gaussian perturbation to the samples and compute weights
\begin{equation}
w_i = \dfrac{q_\flowparams(U_i|C)^{-\beta} p(o | \tau_i)^{\frac{1}{\alpha}}}{ \frac{1}{R} \sum^R_{j=1} q_\flowparams(U_j|C)^{-\beta} p(o| \tau_j)^{\frac{1}{\alpha}}}
\label{eq:update_law}
\end{equation}

\noindent where $p(o|\tau) = \mathrm{exp}(-\cost(\tau))$. These weights represent a trade-off between low-cost and high entropy control sequences controlled by hyperparameters $\alpha$ and $\beta$. The weights and particles $\{U_{1..R}, w_{1..R}\}$ effectively approximate a posterior which is closer to the optimal $q(U|C)$. At each iteration of training, we take one gradient step to maximize the likelihood of $U_{1..R}$ weighted by  $w_{1..R}$, then resample a new set $U_{1..R}$. The flow training loss for this iteration is
\begin{equation}
    \mathcal{L}_{flow} = -\sum^R_{i=1} w_i \log q_\flowparams(U_i|C)
    \label{eq:flow_loss}
\end{equation}
This process is equivalent to performing mirror descent on the variational free energy, see \cite{okada_variational_mpc} for a full derivation. In practice, when sampling $U_{1..R}$ from $q_\flowparams(U|C)$ we add an additional Gaussian perturbation to the samples, decaying the magnitude of the perturbation during training. While this means we are no longer performing gradient descent on $\mathcal{F}$ exactly, we found that this empirically improved exploration during training.
%And the gradient of the loss is
%\begin{equation}
%    \nabla \mathcal{L}_{flow} = -\sum^K_{i=1} w_i \nabla \log %q_\flowparams(U_i|C)
%\end{equation}
%\todo{do we need to put the gradient here, it seems obvious? I would cut it}
To train the parameters of our system we perform the following optimization via stochastic gradient descent:
\begin{equation}
    \min_{\encoderparams, \vaepriorparams, \decoderparams, \contextnetparams, \flowparams} \mathcal{L}_{flow} + a \mathcal{L}_{VAE}
    \label{eq:total_loss}
\end{equation}
for scalar $a \geq 0$. We use a combined loss and train end-to-end so that $h$ is explicitly trained to be used to condition the control sequence posterior. We then continue training the control sequence posterior with a fixed VAE with the following optimization:
\begin{equation}
    \min_{\contextnetparams, \flowparams} \mathcal{L}_{flow}
\end{equation}

\subsection{FlowMPPI}
\label{method:flowmppi}

We present a method for using the learned control sequence posterior for a control task. Given a $\context$ computed from $(x_0, x_G, E)$, the control sequence posterior $q_\flowparams(U|\context)$ can be used as a sampling distribution in a sampling-based MPC approach. We propose a method for using the control sequence posterior with MPPI \cite{mppi}, which we term FlowMPPI (Algorithm \ref{alg:flowmppi}). 

MPPI iteratively perturbs a nominal control sequence with Gaussian noise and performs a weighted sum of the perturbations to find a new control sequence. Empirically, we found that standard MPPI is good at performing local optimization on an already-feasible nominal trajectory. On the other hand, the control sequence posterior is able to sample collision-free goal-directed trajectories, but locally improving trajectories with samples is difficult as small changes in the control sequence posterior latent space $Z$ often lead to large differences in the resulting control sequence. As a result, we observed that we obtained trajectories which reached the goal and avoided obstacles with very few samples, however the cost of the best trajectory did not improve much with more iterations of MPPI. 

FlowMPPI combines sampling in the latent space $Z$, and sampling perturbations to trajectories to get the advantages of both. For a given sampling budget $K$, we generate half of the samples from perturbing the nominal trajectory as in MPPI, and the other half from sampling from the control sequence posterior. These samples will be combined as in standard MPPI. Since the control sequence posterior is invertible w.r.t $U$, a given nominal trajectory $U$ can be transformed to a latent state $Z$. For the samples from the control sequence posterior, we apply a perturbation cost on the distance of the sampled trajectory from the nominal in latent space. This cost mirrors a similar cost in standard MPPI which penalizes perturbations based on distance to the nominal in the control space.
\begin{algorithm}
\caption{A single step of FlowMPPI, this will run every timestep until task is completed or failure is reached. }%}\reducedstate, \reducedcommand, \robotconfig')$}
%\hspace*{\algorithmicindent} 
\textbf{Inputs:} Cost function $J$, previous nominal trajectory $U$, Context vector $C=\contextnet(x_0, x_g, h)$, control sequence posterior flow $f_\flowparams$, MPPI hyperparameters $(\lambda, \Sigma)$,  Horizon T, Samples $K$
\begin{algorithmic}[1]
    \Function{FlowMPPIStep}{}
    \LineComment{Perform shift operation on nominal U}
    \For{t $\in \{1, ..., T-1\}$} 
        \State $U_{t-1} \gets U_{t}$
    \EndFor
    \State $U_{T-1} \sim \mathcal{N}(0, \Sigma)$
    \LineComment{Map nominal controls to $f_\flowparams$ latent space}
    \State $Z \gets f_\flowparams^{-1}(U, C)$
    \LineComment{Generate samples by perturbing nominal U}
    \For{k $\in \{1, ..., \frac{K}{2}\}$}
        \State $\epsilon_U \sim \mathcal{N}(0, \Sigma)$
        \State $U_k \gets U + \epsilon_U$ 
        \State $\tau_k \sim p(\tau|U_k)$ \Comment{Sample trajectory}
        \State $S_k \gets J(\tau_k) + \lambda U_k \Sigma^{-1} \epsilon_U$ \Comment{Compute cost}
    \EndFor
    \LineComment{Generate samples from control sequence posterior}
    \For{k $\in \{\frac{K}{2} + 1, ..., K\}$} 
        \State $\epsilon_Z \sim \mathcal{N}(0, I)$
        \State $U_k \gets f_\flowparams(\epsilon_Z, C)$
        \State $\tau_k \sim p(\tau|U_k)$ \Comment{Sample trajectory}
        \State $S_k \gets J(\tau_k) + \lambda \epsilon_Z (Z - \epsilon_Z)$ \Comment{Compute cost}
    \EndFor
    \LineComment{Compute new nominal U}
    \State $ \beta \gets \min_k S_k$
    \State $\eta = \sum^K_{k=1} \exp (-\frac{1}{\lambda} (S_k - \beta))$
    \For{k $\in \{1, ..., K\}$}
       \State $w_k \gets \frac{1}{\eta} \exp (-\frac{1}{\lambda} (S_k - \beta))$
    \EndFor
    \State $U \gets \sum^K_{k=1} w_k U_k$
    \State \Return $U$
\EndFunction
\end{algorithmic}
\label{alg:flowmppi}
\end{algorithm}

\subsection{Generalizing to OOD Environments} 
A novel environment can be OOD for the control sequence posterior and result in poor performance. %This is because we learn over a dataset of environments; other methods that do not perform learning do not suffer from this issue. 
We present an approach where we \textit{project} the OOD environment embedding $\envembedding$ in-distribution in order to produce low-cost trajectories when it is used as part of the input to $f_\flowparams$. The intuition behind this approach is that our goal is to sample low-cost trajectories in the current environment. Given that $f_\flowparams$ will have been trained over a diverse set of environments, if we can find an in-distribution environment that would elicit similar low-cost trajectories, then we can use this environment as a proxy for the actual environment when sampling from the flow. Thus we avoid the problem of samples from the control sequence posterior being unreliable when the input is OOD.

In order to do this projection, we first need to quantify how far out-of-distribution a given environment is. Once we have such an OOD score, we will find a proxy environment embedding $\hat{\envembedding}$ by optimizing the score, while also regularizing to encourage low-cost trajectories. For the OOD score, we use the VAE we have discussed in section \ref{sec:vae}. VAEs and other deep latent variable models have been used to detect OOD data in prior work \cite{vae_ood1, vae_ood2, ood_gen_models2}, however these methods are typically based on evaluating the likelihood of an input, in our case $p(E)$. For VAEs this requires reconstruction. We would like to avoid using reconstruction in our OOD score for two reasons. First, reconstruction, particularly of a 3D SDF, adds additional computation cost and we would like to evaluate the OOD score in an online control loop. Second, optimizing an OOD score based on reconstruction would drive us to find an environment embedding proxy which is able to approximately reconstruct the \textit{entire} environment. This makes the problem more difficult than is necessary, as we do not need $\hat{\envembedding}$ to accurately represent the entire environment, only to elicit low-cost trajectories from the control sequence posterior.  

\begin{figure}
    \centering
    \includegraphics[width=0.45\textwidth]{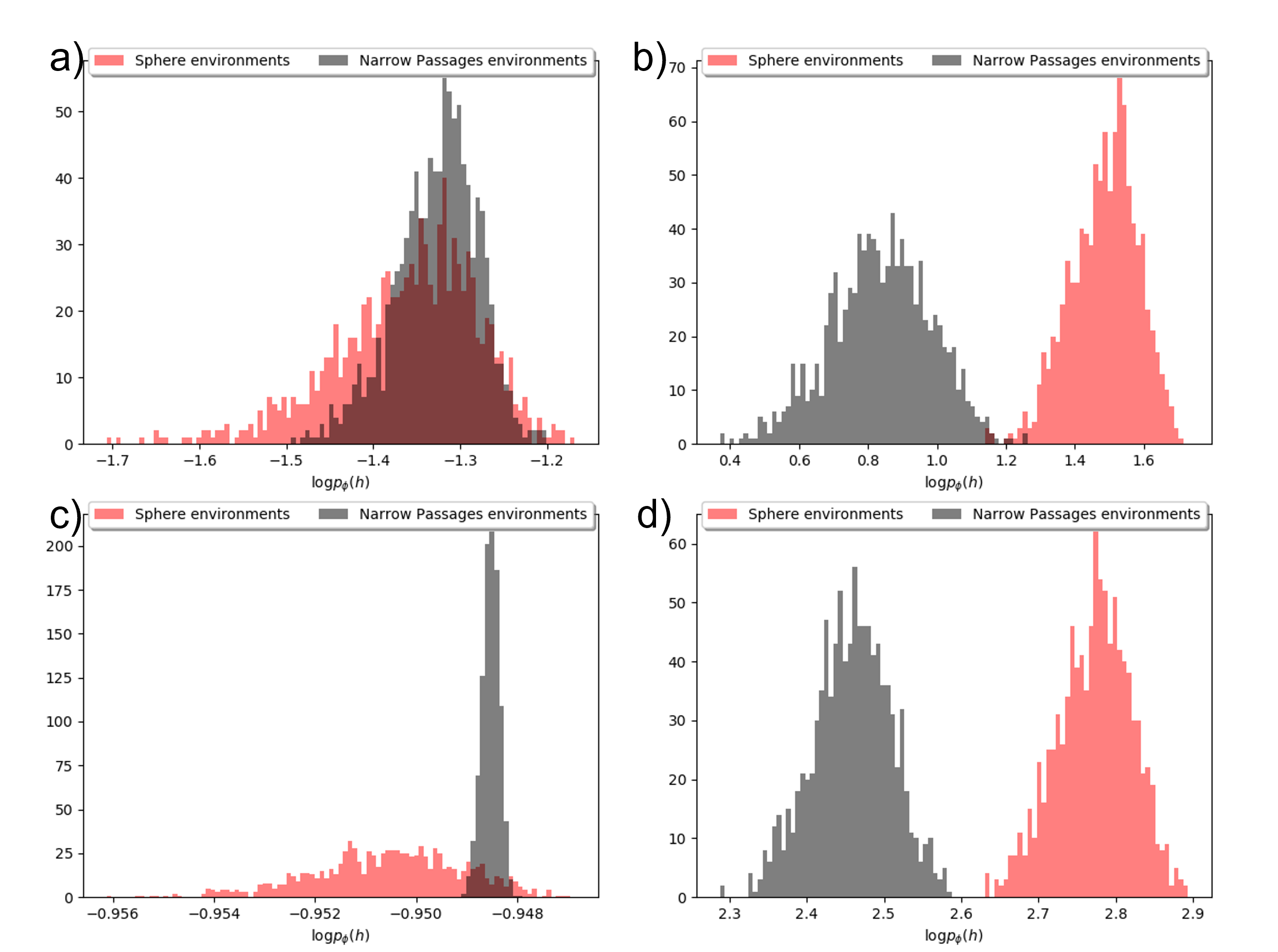}
    \caption{Comparison of our OOD scores with using a VAE with a standard Gaussian prior for in-distribution (red) and out-of-distribution (grey) simulated environments. a) planar navigation using a Gaussian prior, b) planar navigation using a Normalizing flow prior, c) 12DoF quadrotor using a Gaussian prior, d) 12DoF quadrotor using a Normalizing flow prior, These scores are computed by sampling $\envembedding$ from $\encoder$ and evaluating $\log \vaeprior$. The score is normalized by the dimensionality of $\envembedding$. We see that our method, shown in (b) and (d), achieves a clear separation between in-distribution and out-of-distribution environments in both cases.}
    \label{fig:ood_scores}
    \end{figure}

To determine how close $h$ is to being in-distribution, we use the following OOD score:
\begin{equation}
    \mathcal{L}_{OOD}(h) = -\log \vaeprior
\end{equation}

%$\mathcal{L}_{OOD}$ measures whether the representation of the environment used for trajectory generation is in-distribution. 

\noindent where $\vaeprior$ is the learned flow prior for the VAE. The intuition for using this as an OOD score is that this term is minimized for the dataset in $\mathcal{L}_{VAE}$, so we should expect it to be lower for in-distribution data. Using a learned prior was shown to improve density estimation over a Gaussian prior \cite{vae_flow_prior} and we found the learned prior yielded much better OOD detection than using a Gaussian prior, which is the standard VAE prior (see Figure \ref{fig:ood_scores}).

We can perform gradient descent on $\mathcal{L}_{OOD}$ to find $\hat{h}$, thus \textit{projecting} the environment to be in-distribution. Note that without regularization this process will converge to a nearby maximum likelihood solution, which may lose key features of the current environment. Since our aim is to sample low-cost trajectories from the control sequence posterior, we use $\mathcal{L}_{flow}$ as a regularizer for this gradient descent. Our intuition here is that in order to generate low-cost trajectories in the true environment, the projected environment embedding should preserve  important features of the environment relevant for that particular planning query. The new environment embedding is then given by
\begin{equation}
    \hat{\envembedding} = \arg \min_\envembedding \: b \mathcal{L}_{OOD} + \mathcal{L}_{flow}
    \label{eq:projection}
\end{equation}

\noindent for scalar $b > 0$. We project $\envembedding$ to $\hat{\envembedding}$ by minimizing the above by gradient descent. This step is incorporated into FlowMPPI in a version of our method FlowMPPIProject. This version of our method will perform $M$ steps of gradient descent on the above combined loss at initialization, followed by a single step at each iteration of FlowMPPIProject. The algorithm for projection is shown in algorithm \ref{alg:project} in appendix \ref{appendix:algs}.

\section{Evaluation}
In this section, we will evaluate our proposed approaches FlowMPPI \& FlowMPPIProject on two simulated systems; a 2D point robot and a 3D 12DoF quadrotor. For each system, we will train the flow on a dataset of starts, goals and environments and evaluate the performance on environments drawn from the same distribution. In addition, for each system we will test on novel environments that are radically different from those used for training and evaluate the generalization of our approach and the ability of FlowMPPIProject to adapt to these OOD environments. For the 12DoF quadrotor system, we additionally evaluate our method in simulation on two environments generated from real-world data from the 2D-3D-S dataset \cite{3d_dataset}, where our goal is to evaluate if the control sequence posterior, trained on simulated environments, can adapt to real-world environments. 

For our novel environments, we select environments which are difficult for sampling-based MPC techniques. We will use the terms ``in-distribution'' and ``out-of distribution'' for environments for the rest of this section, but note that these terms are relative to the set of environments which we use to train our method. Being out-of-distribution has no bearing on the non-learning based baselines. The performance of non-learning sampling-based MPC algorithms depends only on the given environment, not its relation to other environments.

\subsection{Systems \& Environments} 
\label{eval:tasks}
\begin{figure}
    \centering
    \includegraphics[width=0.45\textwidth]{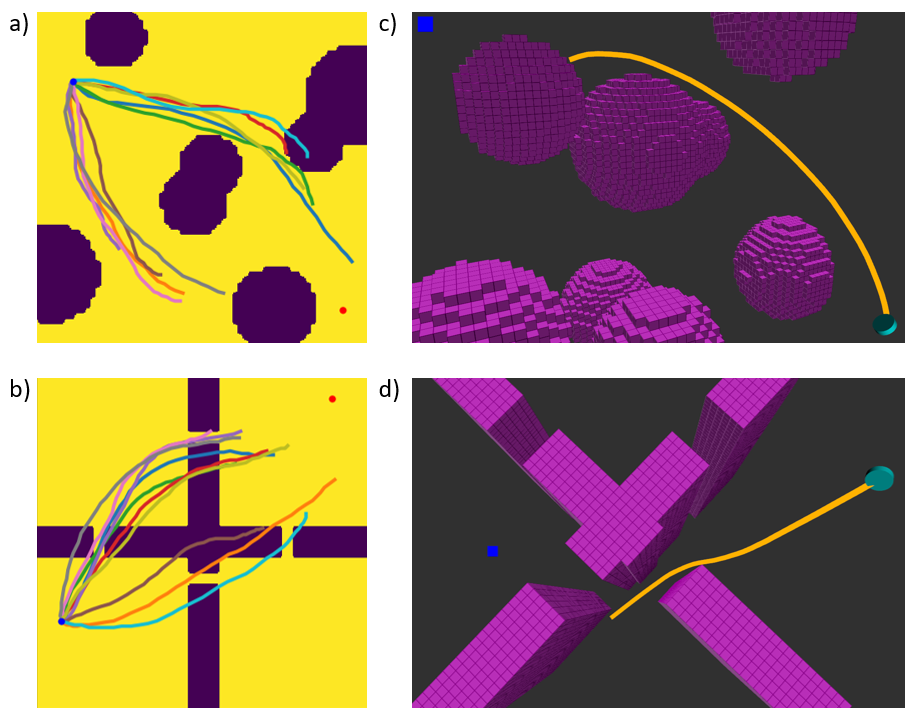}
    \caption{Examples of our 'in-distribution' environments (top) and 'out-of-distribution' environments (bottom). a) The sphere environment for the planar navigation task, showing sampled trajectories from the flow. b) The narrow passages environment for planar navigation, we see that the samples from the flow are goal orientated and generally toward the passages, but most are generally not collision free. c) The sphere environment for the 12DoF quadrotor. d) The narrow passages environment for the 12DoF quadrotor}
    \label{fig:environments}
    \vspace{-0.6cm}
\end{figure}

\begin{figure*}[t]
    \centering
    \includegraphics[width=0.6\textwidth]{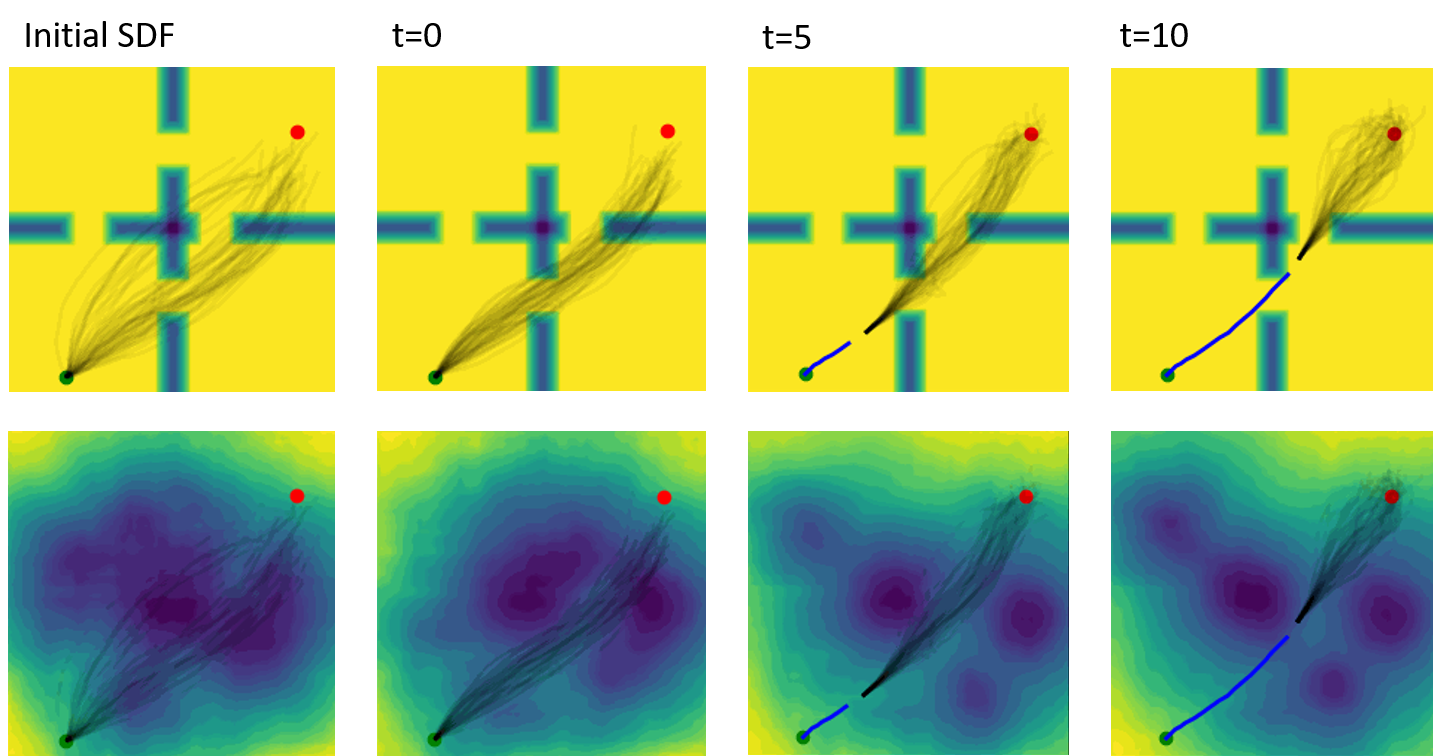}
    \caption{The projection process visualized for the planar navigation task. We visualize the projected environment embedding using the VAE decoder. Note that decoding $\envembedding$ is only used for training the VAE and visualization, it is not necessary for projection. The top shows the environment and sampled trajectories from $q_\flowparams(U|\context)$. The bottom shows the same samples overlaid on a reconstruction of projected environment embedding $\hat{\envembedding}$. On the left, the initial SDF is very poor. As the task progresses, the iterative projection results in an SDF that resembles the training environment more. The environment embedding encodes obstacles that result in a trajectory which traverses the narrow passage. Notice however, that regions that are not relevant for this planning task, such as the top left corner, do not need to accurately represent the environment.}
    \label{fig:projection}
    \vspace{-0.25cm}
\end{figure*}

In this section we will introduce the systems and the environments we use for evaluation. For all systems and environments, a task is considered a failure if there is a collision or if the system does not reach the goal region within a timeout of 100 timesteps. The cost function for both systems is given by $J(\tau) = 100 d_G(x_T) + \sum^{T-1}_{t=1} 10d_G(x_t)_2 + \sum^{T}_{t=1} 10000 D(x_t)$, where $T$ is the MPC horizon, $d_G$ is a distance to goal function, and $D$ is an indicator function which is 1 if $x_t$ is in collision and 0 otherwise. For all of our experiments, the control horizon $T=40$. We use a Gaussian prior over controls is $p(U) = \mathcal{N}(0, \sigma^2 I)$ which induces a cost on the squared magnitude of the actions. For all of our experiments the dynamics are deterministic. Further details of the generation of training data can be found in appendix \ref{appendix:envs}.

\subsubsection{Planar Navigation} The robot in the planar navigation task is a point robot with double-integrator dynamics. The goal is to perform navigation in an environment cluttered with obstacles. The state and control dimensionality are 4 and 2, respectively. The environment is represented as a $64 \times 64$ SDF. Examples of the training and evaluation environments are shown in Figure \ref{fig:environments} (a \& b). The training environments consist disc-shaped obstacles, where the size, location and number of obstacles is randomized. The out-of-distribution environment consists of four rooms, with narrow passages randomly generated between them. The location of the passages is randomized for each OOD environment. The distance to goal function is $d_G(x) = ||x - x_G||_2$. The goal region for this task is given by $\mathcal{X}_G = \{x : ||x - x_G||_2 < 0.1$\}. The dynamics for this system are shown in appendix \ref{appendix:2d_dynamics}. %For collision checking we consider the system to be a point robot.

\subsubsection{3D 12DoF Quadrotor}
This system is a 3D 12DoF underactuated quadrotor with the shape of a short cylinder. It has state and control dimensionality of 12 and 4, respectively. As with the planar navigation task, the goal is to perform navigation in a cluttered environment. Examples of the training and evaluation environments are shown in Figure \ref{fig:environments} (c \& d). The training environment consists of spherical obstacles of random size, location, and number, and the out-of-distribution environment of four rooms separated by randomly generated narrow passages. The environment is represented as a $64 \times 64 \times 64$ SDF. The goal region is specified as a 3D position $p_G$. The distance to goal function is $d_G(x) = ||A x - p_G||_2 + 0.01 ||Bx||_2$ where $A$ selects the position components from the state $x$, and $B$ selects the angular velocity components. The goal region is $\mathcal{X}_G = \{x : d_G(x) < 0.3$\}. We also tested in two simulation environments generated from real-world data (shown in \ref{fig:realworld}). %When evaluating in the real-world environments, we roll out the dynamics in simulation. 
The dynamics for this system are shown in appendix \ref{appendix:quadrotor_dynamics}.

\subsection{OOD Score and Projection}
To confirm the efficacy of our OOD score, we computed this score for the training and OOD environments for each system above. Figure \ref{fig:ood_scores} shows that this score is clearly able to distinguish in-distribution environment embeddings from OOD ones. To show the necessity of using both the OOD score and the regularization in projection, we perform an ablation on these two components in appendix \ref{appendix:ablation} for the quadrotor system.

%For collision checking we model the quadrotor as a disc\todo{technically a disc is infinitely thin, is this what you mean or is it a short cylinder?} and useand collision check 9 spheres which cover the volume of the quadrotor. 

%\subsubsection{7D Manipulator}
%This system is a kinematic 7DoF KUKA LBR iiwa arm, where both the state and control are 7 dimensional; The state is the joint configuration, and the control is a change in the joint configuration. Examples of the training and evaluation environments are shown in Figure \ref{fig:environments} (e \& f). The training environment consists of a tabletop manipulation scene amongst clutter. Rather than a goal state, for this task we specify a goal end effector position $p_{ee}$. The distance to goal function is $d_G(x) = ||\texttt{forwardKinematics}(x) - p_{ee}||_2$. The goal region is $\mathcal{X}_G = \{x : d_G(x) < 0.1$\}. For collision checking we approximate the volume of the arm as a set of 8 spheres. 
%\rev{TODO: add details of the simulated OOD environment for victor}

\subsection{Network Architectures}
\label{eval:architectures}
For both the control sequence posterior flow $\flow$ and the VAE prior $\vaeprior$ we use an architecture based on Real-NVP \cite{real-nvp}. For the VAE prior $\vaeprior$ we use a flow depth of 4, while for the control sampling flow $\flow$ we use a flow depth of 10. For the control sampling flow we use the conditional coupling layers from \cite{conditional_flow}. For the VAE encoder we use four CNN layers with a kernel of 3 and a stride of 2, followed by a fully connected layer. For the VAE decoder we used a fully connected layer followed by four transposed CNN layers. For the 3D case we use 3D convolutions. The dimensionality of both $\envembedding$ and $\context$ was $64$ for the planar navigation environments $256$ for 3D 12DoF quadrotor environments. $\contextnet$ was defined as an MLP with a single hidden layer of size $256$. For nonlinear activations we used ReLU throughout.    

\subsection{Training \& Data} For training, we use $10000$ randomly generated environments for planar navigation task, and $20000$ for the 3D 12DoF quadrotor task. At each epoch, for each environment, we randomly select one of $100$ start and goal pairs. We train the control sequence posterior flow $\flow$, the VAE parameters ($\encoderparams, \vaepriorparams, \decoderparams$) and the context MLP $\contextnet$ end-to-end using Adam for $1000$ epochs with a learning rate of $1e^{-3}$, with a decay rate of $0.9$ every 50 epochs. After 100 epochs, we freeze the VAE and do not continue training with $\mathcal{L_{VAE}}$. This is primarily because the VAE converges quickly and training proceeds more quickly without reconstruction. When training the VAE we divide the loss by the total dimensionality of the SDF and use $a = 5$. For every environment for the flow training, hyperparameters we use $\beta = 1$ and we linearly increase $\alpha$ from $1$ to $500$ during training. Empirically we found that low initial $\alpha$ was required for the flow to learn to generate goal-directed trajectories early on during training, and that increasing $\alpha$ later during training increases the diversity of the flow sampling distribution. A more details list of training hyperparameters can be found in appendix \ref{appendix:training}

\subsection{Baselines}
For our baselines we use several state-of-the-art sampling-based MPC methods: MPPI \cite{mppi}, Stein Variational MPC (SVMPC) \cite{stein_mpc} and improved CEM (iCEM) \cite{iCEM}. 
MPPI uses a Gaussian distribution as the sampling distribution, iCEM uses colored noise, and SVMPC uses a mixture of Gaussians. For each baseline, we tune the hyperparameters to get the best performance based on the training environments, and maintain these hyperparameters when switching to the out-of-distribution environments. When evaluating our two proposed methods and the baselines, each method is given the same sampling budget per timestep. This means that for methods that require multiple iterations per timestep, the sampling budget is distributed across the iterations. A more detailed list of the hyperparameters for each controller can be found in appendix \ref{appendix:controller_params}. Evaluating $\mathcal{L}_{flow}$ during projection requires sampling and evaluating control sequences. When we consider the sampling budget of different algorithms in evaluation, we will include these samples. FlowMPPIProject uses half of the allowed sampling budget during the project step, and the other half for the FlowMPPI control algorithm. While it does take longer to sample from the flow than from the distributions in the baselines, we observe that the cost of evaluating control sequences dominates over the cost of sampling. For example, for the 3D 12DoF quadrotor system, sampling 1024 control sequences from the flow and evaluating the cost of these control sequences takes on average $9$ms and $80$ms, respectively on an i7-8700K CPU and Nvidia 1080 Ti GPU.

\begin{table*}[b]
\begin{center}
\begin{tabular}{ c|c|c|c|c|c|c|c|c|c } 
& & \multicolumn{2}{c|}{In-Distribution} & \multicolumn{6}{|c}{Out-of Distribution} \\ 
& & \multicolumn{2}{c|}{K=512} & \multicolumn{2}{c|}{K=256} & \multicolumn{2}{c|}{K=512} & \multicolumn{2}{c}{K=1024} \\
 System & Controller & Success & Cost & Success & Cost & Success & Cost & Success & Cost \\
 \hline
 \multirow{5}{*}{Planar Navigation} & MPPI & 0.89 & 1925 & 0.19 & 3180 & 0.29 & 2948 & 0.36 & 2840 \\
  & SVMPC & 0.97 & \textbf{1523} & 0.18 & 3032 & 0.22 & 2727 & 0.25 & 2666 \\
  & iCEM & 0.97 & 1531 & 0.46 & \textbf{2467} & 0.59 & \textbf{2145} & 0.62 & 2127 \\
  & FlowMPPI & \textbf{0.99} & 1705 & 0.62 & 2731 & 0.75 & 2155 & 0.84 & 2104 \\
  & FlowMPPIProject & \textbf{0.99} & 1737 & \textbf{0.65} & 2690 & \textbf{0.77} & 2155 & \textbf{0.87} & \textbf{2059}\\
 \hline
 \multirow{5}{*}{12DoF Quadrotor} & MPPI & 0.57 & 3589 & 0.05 & 4809 & 0.11 & 4724 & 0.27 & 4351 \\
  & SVMPC & 0.55 & 3745 & 0.11 & 5588 & 0.21 & 4947 & 0.44 & 4486 \\
  & iCEM & 0.96 & 2724 & 0.35 & 4388 & 0.47 & 4157 & 0.63 & 3795 \\
  & FlowMPPI & 0.92 & 2595 & 0.56 & 3805 & 0.72 & 3601 & 0.84 & 3421 \\
  & FlowMPPIProject & \textbf{0.98} & \textbf{2437} & \textbf{0.71} & \textbf{3688} & \textbf{0.83} & \textbf{3443} & \textbf{0.93} & \textbf{3200}
  \\
  \hline
 %\multirow{5}{*}{7D Manipulator} & MPPI & & & & & & & &\\
 % & SVMPC & & & & & & & &\\
 % & iCEM & & & & & & & &\\
 % & FlowMPPI & & & & & & & &\\
 % & FlowMPPIProject & & & & & & & &\\
 % \hline
\end{tabular}
\caption{Comparison of methods on 100 randomly generated environments, starts and goals for both in distribution and out-of-distribution training environments. Performance on out-of-distribution environments is evaluated for three different sampling budgets. }
\label{table:results_simulated}
\end{center}
\end{table*}

\begin{table}[h]
\begin{center}
\begin{tabular}{ c|c|c|c|c } 
& \multicolumn{2}{c|}{\textbf{Rooms Environment}} & \multicolumn{2}{|c}{\textbf{Stairway Environment}} \\ 
 Method & Success & Cost & Success & Cost \\
 \hline
 MPPI & 0.83 & 3111 & 0.32 & 3019 \\
 SVMPC & 0.68 & 7556 & 0.49 & 2770\\
 iCEM & 0.92 & 2412 & 0.58 & 2623\\
 FlowMPPI & 0.87 & 2375 & 0.5 & 2463 \\
 FlowMPPIProject & \textbf{0.97} & \textbf{1972} & \textbf{0.85} & \textbf{1745} \\
\end{tabular}
\caption{Comparison of methods for the 3D 12DoF quadrotor navigation task with two environments generated from real-world data. The rooms environment is shown in figure \ref{fig:environments} (b) and the stairway environment is shown in figure \ref{fig:environments} (a). We evaluate on 100 randomly sampled starts and goals in each environment.}
\label{table:real_env_results}
\end{center}
\vspace{-0.8cm}
\end{table}
\subsection{Results}
The results comparing our MPC methods to baselines are shown in Tables \ref{table:results_simulated} and \ref{table:real_env_results}. For the planar navigation case, we see that FlowMPPI and FlowMPPIProject are competitive with the baselines on the training environments. Our method reaches the goal region more often, while attaining slightly higher average cost. For the out-of-distribution environments, our method reaches the goal significantly more often. For example, with a sampling budget of $256$, the success rates for FlowMPPIProject is $0.65$ and increases to $0.87$ for a sampling budget of 1024. The next closest baseline, iCEM, has successes rates of $0.46$ and $0.62$ for sampling budgets of $256$ and $1024$, respectively. The projection process for the planar navigation task is visualized for an OOD environment in Figure \ref{fig:projection}.

We observed during this experiment that when iCEM and SVMPC are able to generate a trajectory which reaches the goal region, they are able to locally optimize this trajectory better than FlowMPPI variants, while FlowMPPI is better able to generate sub-optimal trajectories to the goal region.

For the quadrotor system, FlowMPPIProject outperforms all other methods in both cost and success rate across all environments and sampling budgets. With a sampling budget of $256$, FlowMPPIProject attains a $71\%$ success rate compared to $35\%$ by iCEM and $11\%$ by SVMPC for OOD environments. For a sampling budget of $1024$ the success rate of FlowMPPIproject rises to $93\%$ vs. $63\%$ for iCEM. The dynamics of this task make it much more difficult, particularly because stabilizing around the goal is non-trivial. We found that the baselines struggled to find trajectories which both reached and stabilized to the goal, and thus were more susceptible to becoming stuck in local minima. 

Table \ref{table:real_env_results} shows the results when evaluating our method in simulation in two environments generated from real-world data. FlowMPPIProject outperforms all baselines in cost \& success rate, despite only being trained on simulated environments consisting of large spherical obstacles. For the challenging stairway environment, FlowMPPIProject achieves $85\%$ success, while the next closest baseline, iCEM, has $58\%$ success. FlowMPPI achieves only $50\%$ success rate for this task, highlighting the importance of projection for real-world environments. 

\section{Conclusion} 
In this paper we have presented a framework for using a Conditional Normalizing Flow to learn a control sequence sampling distribution for MPC based on the formulation of MPC as Variational Inference. The control sequence posterior samples control sequences which result in low-cost trajectories that avoid collision. We have shown how this control sequence posterior can be used in a sampling-based MPC method FlowMPPI. We have also proposed a method for adapting this control sequence posterior to OOD environments by \textit{projecting} the representation of the environment to be in-distribution, essentially ``hallucinating'' an in-distribution environment which elicits low-cost trajectories from the control sequence posterior. We have demonstrated that our proposed MPC methods FlowMPPI and FlowMPPIProject offer large improvements over baselines in difficult environments, and that by performing the environment projection we can successfully transfer a control sequence posterior learned with simulated environments to environments generated from real-world data.
\label{sec:conclusion}

\section*{Acknowledgments}
This work was supported in part by NSF grants IIS-1750489 and IIS-2113401, and ONR grant N00014-21-1-2118. We would like to thank the other members of the Autonomous Robotic Manipulation Lab at the University of Michigan for their insightful discussions and feedback.

%% Use plainnat to work nicely with natbib. 
{\small
\bibliographystyle{plainnat}
\bibliography{references}
}
\clearpage

\appendix
\subsection{Variational Inference for Finite-Horizon Stochastic Optimal Control}
\label{appendix:vimpc}
The variational posterior over trajectories is defined by the dynamics and the variational posterior over actions:
\begin{align}
\begin{split}
    q(\tau |x_0) &= q(X,U |x_0) \\
    &=  p(X|U, x_0)q(U) \\
    &= q(U)\prod^T_{t=0} p(x_{t+1} | x_t, u_t)
\end{split}
\end{align}
We will omit the dependence on the initial state $x_0$ for convenience.
\begin{align}
\begin{split}
&\mathcal{KL}\left(q (\tau) || p(\tau | o=1) \right) = \int q(\tau) \log \frac{q(\tau)}{p(\tau | o=1)} d\tau \\
&= \int q(X, U) \log \frac{p(X|U)q(U)p(o=1)}{p(o=1|X, U)p(X|U)p(U)}dXdU 
\end{split}
\end{align}
Since $p(o=1)$ on the numerator does not depend on U, when we minimize the above divergence it can be dropped. The result is minimizing the below quantity, the \textit{variational free energy} $\mathcal{F}$. 
\begin{align}
\mathcal{F} &= \int q(X, U) \log \frac{q(U)}{p(o=1|X, U)p(U)}dXdU \\
\begin{split}
&= -\mathbb{E}_{q(X, U)} \left[ \log p(o|X, U) + \log p(U) - \log q(U)\right] \\ 
\end{split} \\
&= \mathbb{E}_{q(X, U)} \left[J(X, U)\right] + \mathcal{KL}(q(U) || p(U)) \\
&= \mathbb{E}_{q(X, U)} \left[\hat{J}(X, U) + \log q(U) \right] \label{app:objective}
\end{align}

For the last two expressions we have used our formulation that the $p(o=1 | X, U) = \exp (-J(X,U))$, where $J$ is the trajectory cost, and we have incorporated the deviation from the prior into the cost function. For example, a zero-mean Gaussian prior on the controls can be equivalently expressed as a squared cost on the magnitude of the controls. 
\subsection{Training \& Architecture Details}
\begin{figure}[H]
    \centering
    \includegraphics[width=0.48\textwidth]{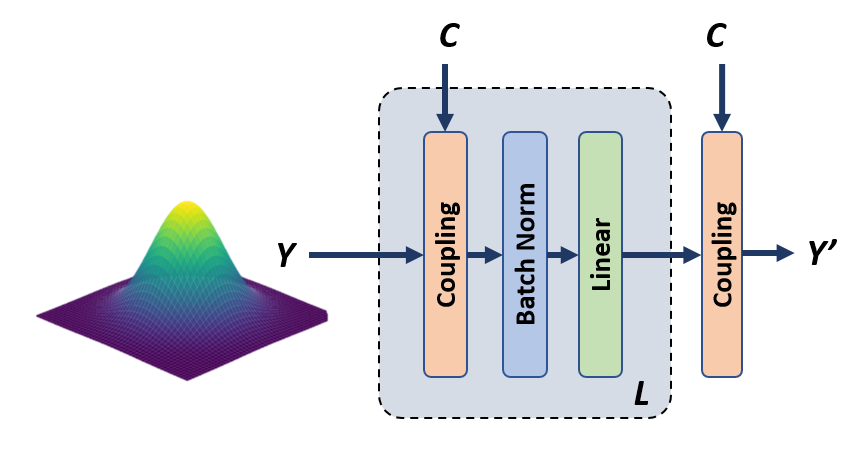}
    \caption{The architecture for both the prior flow and the control sequence posterior flow, based on \cite{real-nvp} and \cite{conditional_flow}, showing a mapping from arbitrary Y to Y'. Each flow consists of L chained transformation blocks. A transformation block consists of a conditional coupling layer, a batch norm layer, and a linear layer. There is a final conditional coupling layer on the output. For the vae prior, there is no context therefore we use standard coupling layers and not conditional coupling layers. }
    \label{fig:flow_architectures}
\end{figure}

\label{appendix:training}
\begin{table}[H]
\begin{center}
\begin{tabular}{ c|c|c } 
 Variable & Planar Navigation & 3D 12DoF Quadrotor \\
 \hline
  control peturbation $\Sigma_e$ & \multicolumn{2}{c}{$1 -  \dfrac{\text{epoch}}{\# \text{epochs}}$} \\
 $\alpha$ & \multicolumn{2}{c}{$500 \dfrac{\text{epoch}}{\# \text{epochs}}$} \\
 $\beta$ & 1 & 1 \\
 $\#$ epochs & 1000 & 1000  \\
 Initial learning rate & $1 \times 10^{-3}$ & $1 \times 10^{-3}$  \\
 Learning rate decay & \multicolumn{2}{c}{$0.9$ every 50 epochs}  \\ 
 $\#$ Training environments & 10000 & 20000 \\
 $\#$ $(x_0, x_G)$ per training env. & 100 & 100 \\ 
 $\envembedding$ dim & 64 & 256 \\
 $a$ & 5 & 5 \\ 
 $b$ & $\dfrac{1}{64}$ & $\frac{1}{1024} $\\
 VAE training epochs & 100 & 100 \\
 $\vaeprior$ flow depth L & 4 & 4 \\
 $f_\flowparams$ flow depth L & 10 & 10 
\end{tabular}
\caption{Training and architecture hyperparameters for each experiment.}
\label{table:training_hyperparams}
\end{center}
\vspace{-0.8cm}
\end{table}

\subsubsection{Hyperparameter Tuning}
There are several hyperparameters to tune in our approach. The scalar $a$ in equation \ref{eq:total_loss} was tuned so that $a \mathcal{L}_{VAE}$ and $\mathcal{L}_{flow}$ were of approximately similar magnitude.
The scalar $b$ in equation \ref{eq:projection} was selected to be equal to the dimensionality of the SDF observation divided by the dimensionality of the latent environment embedding. This value was chosen initially to make the projection loss similar across the quadcopter and the double integrator, and we found this automatic tuning worked well in practice. Hyperparameters $\alpha, \beta$ together control the trade-off between entropy and optimality. We kept $\beta$ fixed and tuned only $\alpha$. To tune $\alpha$, for each experiment we performed a grid search and selected the value of $\alpha$ that resulted in the best performance in the training environment when used with FlowMPPI.

\subsection{Environment details}
\label{appendix:envs}
The environments are $4m \times 4m$, and generated as occupancy grids, from which we compute the SDF. For each training environment, we randomly sample 100 start \& goal pairs such that they are always collision free, and within the bounds of the voxel grid. We sample start velocities from a Normal distribution, and set the goal velocity to be zero. During evaluation, for both the in-distribution and out-of-distribution environments, we sample $100$ start, goal and environment tuples and evaluate all methods on these tuples. The exception to this is the real-world environments, where we keep the environments fixed and sample $100$ start and goal pairs per real-world environment and evaluate all methods on these pairs. To ensure the navigation problem is non-trivial, we sample starts and goals that are at least $4 m$ away.
\subsubsection{Real-world environments}
The two real world environments are taken from area 3 from the 2D-3D-S dataset \cite{3d_dataset}. To generate the two environments, we used the 3D mesh from the dataset and defined a subset of the area to be the environment. We then generated an occupancy grid by densely sampling the mesh, which we then used to compute the SDF. 
\subsubsection{Planar Navigation}
\label{appendix:2d_dynamics}
The dynamics for the planar navigation system are
\begin{equation}
    \begin{bmatrix}
    x \\
    y \\
    \dot{x} \\
    \dot{y} 
    \end{bmatrix}_{t+1} = 
    \begin{bmatrix}
    1 & 0 & \Delta t & 0 \\
    0 & 1 & 0 & \Delta t \\
    0 & 0 & 0.95 & 0 \\
    0 & 0 & 0 & 0.95 
    \end{bmatrix}
    \begin{bmatrix}
    x \\
    y \\
    \dot{x} \\
    \dot{y} 
    \end{bmatrix}_t + 
    \begin{bmatrix}
    0 & 0  \\
    0 & 0 \\
    \Delta t & 0 \\
    0 & \Delta t \\
    \end{bmatrix} \mathbf{u}
\end{equation}

\subsubsection{12DoF Quadrotor}
\label{appendix:quadrotor_dynamics}
The dynamics for the 12DoF quadrotor are from \citet{quadrotor_dynamics} and are given by
\begin{equation}
    \begin{bmatrix}
    x \\
    y \\
    z \\
    p \\
    q \\
    r \\
    \dot{x} \\
    \dot{y} \\
    \dot{z} \\
    \dot{p} \\
    \dot{q} \\
    \dot{r} \\
    \end{bmatrix}_{t+1} = 
    \begin{bmatrix}
    x \\
    y \\
    z \\
    p \\
    q \\
    r \\
    \dot{x} \\
    \dot{y} \\
    \dot{z} \\
    \dot{p} \\
    \dot{q} \\
    \dot{r} \\
    \end{bmatrix}_t + \Delta t
    \begin{bmatrix}
    \dot{x} \\
    \dot{y} \\
    \dot{z} \\
    \dot{p} + \dot{q} s(p) t(q) + \dot{r} c(p) t(q) \\
    \dot{q} c(p)  - \dot{r} s \dot{p} \\
    \dot{q} \dfrac{s(p)}{c(q)} + \dot{r} \dfrac{c(p))}{c(q)} \\
    - ( s (p) s (r) + c(r) c(p) s(q)) K \frac{u_1}{m} \\
    - (c(r) s(p) - c(p) s(r) s (q)) K \frac{u_1}{m} \\
    g - c(p) s(q)) K \frac{u_1}{m} \\
    \frac{(I_y - I_z) \dot{q} \dot{r} + K u_2}{I_x} \\
    \frac{(I_z - I_x) \dot{p} \dot{r} + K u_3}{I_y} \\
    \frac{(I_x - I_y) \dot{p} \dot{q} + K u_4}{I_z} \\
    \end{bmatrix}_t
\end{equation}
Where $c(p), s(p), t(p)$ are $\cos, \sin, \tan$ functions respectively. We use a parameters $m=1, I_x=0.5, I_y=0.1, I_z=0.3, K=5, g=-9.81$. The quadrotor geometry is modeled as a cylinder with radius $0.1m$ and height $0.05m$. 
\subsection{Controller details}
\label{appendix:controller_params}
\begin{table}[H]
\begin{center}
\begin{tabular}{ c|c|c } 
 Variable & Planar Navigation & 12DoF Quadrotor \\
 \hline
 Control Horizon $H$ & 40 & 40 \\
 Trial length $T$ & 100 & 100 \\
 Control prior $\sigma$ & 1 & 4 \\
 Dynamics $\Delta t$ & 0.05 & 0.025
\end{tabular}
\caption{Controller agnostic parameters used for the evaluations}. 
\label{table:controller_general_hyperparams}
\end{center}
\end{table}

\begin{table}[H]
\begin{center}
\begin{tabular}{ c|c|c|c } 
 Controller & Variable & Planar Navigation & 12DoF Quadrotor \\
 \hline
 \multirow{3}{*}{MPPI} & $\lambda$ & 1 & 1 \\
 & $\Sigma$ & 0.9 & 0.5 \\
 & iterations & 1 & 4\\ 
 \hline
 \multirow{5}{*}{SVMPC} & $\Sigma$ & 1 & 0.5 \\
 & $\#$ particles & 4 & 4 \\
 & Learning rate & 1 & 0.5 \\
 & iterations & 4 & 4 \\
 & warm-up iterations & 25 & 25 \\
 \hline
 \multirow{6}{*}{iCEM} & $\Sigma$ & 0.75 & 0.5 \\
 & noise parameter & 2.5 & 3 \\
 & $\%$ elites  & 0.1 & 0.1 \\
 & $\%$ kept elites & 0.3 & 0.5 \\
 & iterations & 4 & 4 \\
 & momentum & 0.1 & 0.1 \\
 \hline
 \multirow{5}{*}{FlowMPPI} & $\lambda$ & 1 & 1 \\
 & $\Sigma$ & 1 & 0.75 \\
 & iterations & 1 & 2 \\
 & $M$ & 10 & 10 \\
 & Proj. learn. rate & $1\times 10^{-2} $& 1$\times 10^{-2}$ \\
\end{tabular}
\caption{Controller hyperparameters used for the experiments for both our proposed method and the baselines.}
\label{table:controller_hyperparams}
\end{center}
\end{table}

\subsection{Algorithms}
\label{appendix:algs}
%\begin{minipage}[T]{\linewidth}
\begin{algorithm}[H]
\caption{Sample from Control Sequence Posterior with Perturbation}%}\reducedstate, \reducedcommand, \robotconfig')$}
%\hspace*{\algorithmicindent} 
\begin{algorithmic}[1]
\Function{SamplePertU}{$\context, \Sigma_{\epsilon}, K$}
\For{i $\in \{k, ..., K\}$}
    \State $Z_k \sim \mathcal{N}(0, I)$
    \State $\epsilon_k \sim \mathcal{N}(0, \Sigma_{\epsilon})$
    \State $U_k \gets f_\flowparams(Z_k, C) + \epsilon_k$
    \State $\hat{Z}_k \gets f_\flowparams^{-1}(U_k, C)$
    \State $q_\flowparams(U_k|C) \gets \text{from $\hat{Z}_k$ via eq. (\ref{background:flow_ll}})$
\EndFor
\State \Return{$\{U_k, q_\flowparams(U_k|C)\}^K_{k=1}$}
\EndFunction
\end{algorithmic}
\label{alg:sample}
\end{algorithm}
%\end{minipage}
%\end{wrapfigure}

\begin{algorithm}[H]
\caption{Flow Training}%}\reducedstate, \reducedcommand, \robotconfig')$}
%\hspace*{\algorithmicindent} 
\textbf{Inputs:} N iterations, K samples, $\Theta^1 = \{\encoderparams^1, \decoderparams^1, \vaepriorparams^1, \contextnetparams^1, \flowparams^1\}$ initial parameters, control perturbation covariance $\Sigma_{\epsilon}$, learning rate $\eta$, loss hyperparameters $(\alpha, \beta)$
\begin{algorithmic}[1]
\For{n $\in \{1, ..., N\}$}
    \State $\envembedding \gets \encoder$
    \State $\hat{E} \gets \decoder$
    \State Compute $\log \vaeprior$  via eq. (\ref{background:flow_ll})
    \State Compute $\mathcal{L}_{VAE}$
    \State $\context \gets \contextnet(x_0, x_G, h)$
    \State $\{U_k, q_\flowparams(U_k|C)\}^K_{k=1} \gets \textproc{SamplePertU}(C, \Sigma_\epsilon, K)$
    \State $\mathcal{L} \gets \mathcal{L}_{VAE}$
    \For{k $\in \{1, ..., K\}$}
        \State $w_k \gets \text{from $(\{U_i, \log q_\flowparams(U_i|C)\}^K_{i=1}, \alpha, \beta\})$ via (\ref{eq:update_law}})$
        \State $\mathcal{L} \gets  \mathcal{L} - w_k \cdot \log q_\flowparams(U_k|C)$
    \EndFor
    \State $\Theta^{n+1} \gets \Theta^{n} - \eta \frac{\partial \mathcal{L}}{\partial \Theta}$
\EndFor
\end{algorithmic}
\label{alg:train}
\end{algorithm}
%\end{minipage}
%\end{wrapfigure}
\begin{algorithm}[H]
\caption{Projection}%}\reducedstate, \reducedcommand, \robotconfig')$}
%\hspace*{\algorithmicindent} 
\textbf{Inputs:} N iterations, K samples, $\encoderparams, \vaepriorparams, \contextnetparams, \flowparams$ parameters, control perturbation covariance $\Sigma_{\epsilon}$, learning rate $\eta$, loss hyperparameters $(\alpha, \beta)$
\begin{algorithmic}[1]
\State $\envembedding^1 \gets \encoder$
\For{n $\in \{1, ..., N\}$}
    \State Compute $\log p_{\vaepriorparams}(h^n)$  via eq. (\ref{background:flow_ll})
    \State $\context \gets \contextnet(x_0, x_G, \envembedding^n)$
    \State $\{U_k, q_\flowparams(U_k|C)\}^K_{k=1} \gets \textproc{SamplePertU}(C, \Sigma_\epsilon, K)$
    \State $\mathcal{L} \gets -p_{\vaepriorparams}(h^n)$
    \For{k $\in \{1, ..., K\}$}
        \State $w_k \gets \text{from $(\{U_i, \log q_\flowparams(U_i|C)\}^K_{i=1}, \alpha, \beta\})$ via (\ref{eq:update_law}})$
        \State $\mathcal{L} \gets  \mathcal{L} - w_k \cdot \log q_\flowparams(U_k|C)$
    \EndFor
    \State $\envembedding^{n+1} \gets \envembedding^{n} - \eta \frac{\partial \mathcal{L}}{\partial \envembedding}$
\EndFor
\end{algorithmic}
\label{alg:project}
\end{algorithm}
%\end{minipage}
%\end{wrapfigure}

\subsection{Additional Results}
\label{appendix:ablation}
\begin{table}[H]
\begin{center}
\begin{tabular}{ c|c|c|c|c|c|c } 
& \multicolumn{2}{c|}{K=256} & \multicolumn{2}{c|}{K=512} & \multicolumn{2}{c}{K=1024} \\
Projection loss& Success & Cost & Success & Cost & Success & Cost \\
 \hline
$\mathcal{L}_{OOD} + \mathcal{L}_{flow}$& \textbf{0.71} & \textbf{3688} & \textbf{0.83} & \textbf{3443} & \textbf{0.93} & \textbf{3200}\\
$\mathcal{L}_{OOD}$ &  0.52 & 3859 & 0.63 & 3704 & 0.89 & 3371 \\
$\mathcal{L}_{flow}$ &  0.6 & 3758 & 0.72 & 3489 & 0.87 & 3226\\
\end{tabular}
\caption{Ablation of the different loss terms in FlowMPPIProject for different sampling budgets for the 12DoF quadrotor out-of-distribution environment}
\label{table:quadrotor_ablation}
\end{center}
\end{table}
\end{document}